# The universal evolutionary computer based on super-recursive algorithms of evolvability

Darko Roglic
Evolutionary Computation Research
Center
Poljudsko set.4, 21000 Split, Croatia

droglic@inet.hr

## ABSTRACT
The bacteria are great champions of evolutionary success. They can survive in the boiling water full of sulfer and salt, in the ice, the high exposures to desiccation and dehydration, under the very high pressure in the sea depths, in the cooling water of nuclear reactors, they have become resistant to the entire arsenal of human made antibiotics and they have even survived experimentaly achieved meteorite blow. In comparison, our computers are fragile monoliths that crash at the slightest hint of an execeptional circumstance. It isn't hard to imagine what a breakthrough it will be in the field of artificial intelligence and artificial life having computational devices with such bacterial efficency.
This work exposes which mechanisms and procesess in the Nature of evolution compute a function not computable by Turing machine. The computer with intelligence that is not higher than one bacteria population could have, but with efficency to solve the problems that are non-computable by Turing machine is represented. This theoretical construction is called Universal Evolutinary Computer and it is based on the superecursive algorithms of evolvability.

## Categories and Subject Descriptors
F1.1. [**Theory of Computation**]: Models of Computation – *Computability theory*
H1.1. [**Information Systems**]: Systems and Information Theory – *General systems theory, Information theory, Value of information*

## 1. INTRODUCTION
Everything that is worth understanding about complex systems, how they behave, how they break down, how they work, can be understood in terms of how they process information. The general hypothesis of this work considers evolution as the universal set of algorithms of how the nature processes informations through all the levels of individual living systems and among them in the course of their survival, growth and development. It allows to introduce the concept of information processing system that I have called the Universal Evolutionary Computer. From mathematical point of view, computers function under the control of algorithms so, to understand and explore the possibilities of computers and their boundaries, we have to study algorithms. The current approaches of artificial evolution (evolutionary programming, genetic algorithms, evolutionary strategies and genetic programming) tipically drive relentlessly to an apriori, invariant objective and stop once they get there. [1] This work introduces evolutionary computation based on superrecursive algorithms that give result without stopping.

Several discoveries in the last few decades represent some fascinating life strategies that bacteria use for managing the conflicting goals – adaptation to unpredictable changes and challenges that require flexibility and variability and the mainteance of already evolved complex systems that require stability. This management is essential for escaping two lethal extremes: the evolutionary death due to insufficient variability and immediate death due to excessive change (error catastrophe). It seems that this stability-variability dualism is inherent to all evolving information systems from simple viruses to the human society. Yet, is it possible to simulate the same inherency in our computers and networks?  Right now, neither of this goals is reachable by conventional Turing machines based on recursive algorithms.

The bacteria offered us some remarkable lessons in evolution. So, do we have the knowledge on our disposal in worldwide scientific and engineering community to make a starting point to transfer analogue world of evolution into the digital one toward computational evolution (the term suggested in 1) rather than into *intelligently designed* artificial evolution? However, our intuitive notion of any biological evolutionary computer with its genetic algorithms has to be captured in a rigorous definiton of Church-Turing thesis (CTT). It claimes that any mathematical model of an algorithm is functionally equivalent to a Turing machine (**TM**).

For many years Church, Turing and many other scientists have spent a great deal of time gathering evidence for CTT and no evidence to the contrary has been found. Yet, it turned out that utilization of more powerfull algorithms was inefficient because they were not differentiated from recursive algorithms. In that sense, Burgin has suggested new paradigmal theory of superrecusive algorithms.[2] One of the central dogma for algorithms which states that an algorithm has to stop when it gives result, has been abandoned. The main restriction that hindered the development of computer prevailed. The superrecursive algorithms are based on a new paradigm for computation that changes computational procedures. As Burgin presented, superrecursive algorithms can compute what has been considered as noncomputable. Now practice has to catch up with the theory and it is urgent to know how to bridge the existing gap. Yet, specific processes in the Nature more powerfull than CTT remain undiscovered.

In this paper specific molecular mechanisms of evolutionary adaptation are proposed and described as algorithms that could pose a challenge to the strong CTT. As we could find, so called SOS–repair mechanism and mismatch repair system (MRS) are individually as well as mutually, among the most comprehensive and representative processes in the Nature of evolution that compute a function not computable by Turing machine. Additional mechanisms of strong untargeted mutations (so called mutators) as well as the model of targeted mutations and recombinations (hypermutable genes) have developed and incorporate into the presented model of evolutionary computer for real applications. Additionally, it has been found that evolvability models overlap with the features of superrecursive models of computation. The evolution implicates what superrecursive theory postulates.

Now we could harness evolutionary processes to help us perform new computations which could not be performed before. To overhaul the definition of computability in the direction of superrecursive computational evolvability Universal Evolutionary Computer (UEC) is suggested.

## 2. THE SUPER-RECURSIVE PROACTIVE WORKING MODE OF EVOLUTIONARY ADAPTATION

Two specific mechanism have been extracted from the molecular biology of bacteria and modeled for algorithms of evolutionary computation: SOS repair (or SOS replication, SOS response, SOS regulon) and DNA mismatch-repair system (MRS), particulary long patch (LPMR) and short patch mismatch repair(SPMR). The thorough knowledge of how these mechanisms work allows to treat them as algorithms. Aditionally, specific carriers of genetically controlled processes of mutation and recombination have been considered: genetic mutators and/or hyper-rec mutants (evident at the level of bacterial populations), inducible mutators and recombinators of activity (of individual cells), hypermutable genes and recombinational hotspots (particular sequences). They are treated theoretically as the patterns of information processes and as the carriers of superrecursive features. They are modeled for evolutionary computation and labeled as $EC_{sos}$, $EC_{mrs}$, $EC_{hm}$, $EC_m$.

Each of these evolutionary information processing systems in general or model for evolutionary computation in particular poses a challenge to the strong CTT. They are suggesting that there are efficiently soluble problems which cannot be efficiently solved on a deterministic Turing machine. They are presenting, individually as well as mutually, emergent nature of evolutionary processes. These models are algorithmic but the superrecursive algorithms provide much more adequate models than recursive algorithms. Consenquently, it is possible to compare independently developed bio-inspired models of evolutionary computer (**EC**) with the model of superrecursive algorithms as the inductive Turing machine. Specifically, inductive Turing machine with structured memory (**smITM**) presented in [2], developed by Burgin, is the closest to the recursive algorithms and the most powerful among ITM. The comparison includes specifics of structure, computational processes and results designation that underline the basic features of information processing system of both models as in the Table 1.

The SOS repair alone is considered here as one of the most representative processes that wholly expresses superrecursive features of evolution. It is the basic repair system of our genes discovered and named by Radman.[3,4] In November 1970 he stunned his coleagues with such a heretical proposal. The dogma of molecular and evolutionary biology was that mutation is an unvoidable stochastic event due to the limits of the precision of biological processes - a trade-off between the ideal and the real. Quite opposite, Radman suspected that bacteria harbor a well-regulated genetically programmed process and through this program bacteria can crank up their mutation rates in stressful situations, helping accelerate their own evolution and rapidly adapt to almost any new situation.

Any unrepaired lesion can stop the regular DNA replication machinery, a potentially lethal event, but, the class of SOS polymerases, allows completion of DNA replication despite lesions in the template. Metaphorically, one could say that some specific SOS genes have the basic strategic instruction: *do not stop with replication*! Actually, when bacteria undergo genotoxic or metabolic stress it is better to incorporate a base that is probably wrong and let replication proceed than not to replicate at all. There is no information by which it is possible to choose the correct base to repair a DNA molecule instantly, so ironically it appeared that DNA repair is the *cause* of the mutation. Such mutator activities are inducable and they are turned on only under strong selective pressure. As soon as growth conditions are restored mutator and hyper-recombination activites could be repressed (by LexA repressor).

Damages, or lesions, in DNA templates that are non-instructive for base pairing interrupt the DNA replication process by blocking the elongation of nascent chains. Translesion synthesis (TLS) is the DNA-copying process that can overcome such blockage and allow the completion of DNA despite the presence of lesions. Even more, the same mechanism starts to generate mutations (Radman called it mutases), hence the system functions merely to generate diversity (as for its own sake) without serving any obvious repair function. From the aspect of information processing and computation this particular process presents transition from terminating computation to intrinsically emerging computation. As we already know, this specific feature is inherent in inductive Turing machine. [2] **ITM** presents the same transition as well as the transition from recursive mode of computation into the superrecursive one.

To show that SOS repair computation is more powerfull than ordinary Turing machine, we need to find a problem solvable by SOS evolutionary computation ($\mathbf{EC_{sos}}$) and insolvable by Turing machine (**TM**). This is the halting problem for an arbitrary TM. Turing proved that no TM can solve this problems for all TMs. It is important to note that SOS repair acts as a postreplicative and inducible mechanism. Accordingly, we can reduce the complex SOS machinery just on the most simple TLS working process with its input and output and it is labeled here as simple $\mathbf{sEC_{sos}}$. TM with three tapes: input, working and output has the same structure. Both, inducible EC and ordinary (recursive) TM, perform similiar steps of computation. The difference is in output. A TM produces a result only when it halts. A $\mathbf{sEC_{sos}}$ produces its result without stopping and it gives a result in a finite time.

In the case of $\mathbf{sEC_{sos}}$ functioning we can consider that it contains an universal Turing machine **U** as a subroutine. Given a word *u* and description D(**TM**) of a Turing machine **TM,** evolutionary machine **sEC$_{sos}$** uses machine **U** to simulate **TM** with the input *u*. While **U** simulates **TM**, the **sEC$_{sos}$** produces 0 on the output tape.This functioning simulates normal growth conditions when some of the SOS genes are expressed at certain levels even in the repressed state according to the affinity of LexA repressor to their SOS box. If machine **U** stops, and this means that **TM** halts being applied to *u*, the **sEC$_{sos}$** produces 1 on the output tape. This functioning simulates activation of SOS genes. (The SOS uses Rec A protein. Stimulated by single-stranded DNA, Rec A is involved in the activation of the LexA repressor thereby inducing the error-prone SOS respons.) According to definition, the result of **sEC$_{sos}$** is equal to 1 when **TM** halts and result of **sEC$_{sos}$** is equal to 0 when **TM** never halts. In such a way **sEC$_{sos}$** solves the halting problem.

The Turing machine can efficiently simulate replication process but for simulation of emergent processes of adaptation the powerful superrecursive models are more adequate. However, even the simple models like **sEC$_{sos}$** and previously introduced theoretical model, **sITM** in [2], can be more powerful than conventional TM. At the same time, the development of their structure allowed models of inductive computation to achieve much higher computing power than those described above. This is in contrast to such a property of conventional Turing machine that by changing its stucture we cannot get greater computing power. [2] That is important for the computational exploitation of SOS repair to its full extent. Simple forms do not allow us to compute much more than conventional TM.

It turned out that the attractive vision of Turing machine was partially wrong, at least on the long term basis. It seems that it has tacitly embodied the idea of perfection. In the schemes of simple operations there was no place for program mistakes or any needs for "problem removing". In reality the vast majority of written programs contain errors. Under the assumption of CTT it is not possible to find a procedure or to write a program that allows us to debug all computer programs. The debugging process is not in the program nature of TM. Furthermore, according to CTT thesis to give a result at all the program has to halt, so one kind of program mistakes is that the program never halts. Apparently, evolutionary program of the Nature leaves this theory in its orthodoxy and the superrecursivity appeared to be much more plausible theory. According to unpredictable changes and challenges, evolution works in the direction toward variability and generating diversity. It favours not only evolutionary adaptation process but it evolves the mechanisms that improve the governing of the adaptation processes themselves. The system possibly appeared more adaptable and/or evolvable.

The molecular mechanism of SOS induction has been known for many years but precise chemistry of mutagenic TLS mechanism in bacteria remaind elusive. The major breakthrough came in 1999 when three laboratories came up with the discovery of a new family of unorthodoxical DNA-synthesizing enzymes, or polymerases, apparently, as ultimate proof of SOS. They are undoubtedly showing that evolution has invested in the development and improvement of a mechanism that generates variability. The two key SOS induced genes required for induced mutagenesis (umuC and dinB) are encoding mutagenic lesion-bypass DNA polymerases. The expression of the *E.coli* DNA polymerases pol V and pol IV increases in response to DNA damage.[4 and reference therein] The former is "error free" polymerases that correctly copies the damaged site (amounting to a memory of the sequence before it was damaged by UV light) so, as the mutator-generating TLS, it is responsible for targeted mutations. The latter is "error-prone" polymerases that erroneously copies the intact DNA so, it is responsible for "untargeted" mutations and consequently it exposes to mismatch repair system.[4 and reference therein] The former activity helps survival and both appear to increase genetic variability.

The both DNA polymerases IV and V add nucleotides one by one in an error-prone fashion: pol V produces base substitution and frameshift mutations opposite the damage site, and pol IV produces predominantly frameshift mutations explained by its capacity to extend synthesis on unpaired termini. [4,5] The slow, step-by-step distributive copying mechanism of all TLS polymerase (II, IV and V) may itself assure that only the troubled sites in DNA are their substrates. As soon as the normal base pairs are created, regular faithful and processive replicative polymerases would take over the replication process.[4,6] According to this functiong advanced TLS model of SOS repair, **EC$_{sos}$** , has developed. As we could find all TLS polymerases belong to the class of the superrecursive algorihtms. The result appeared in finite time and without stopping of replication process. The normal base pair created by TLS process could be treated as equivalent to the output word *w* of inductive computation that is not changed. Consequently, the SOS repair computation uses the most direct way to determine a result when the algorithm does not stop functioning. So, it possible to say that inductive SOS computation by **EC$_{sos}$** are the superrecursive algorithms closest to the recursive algorithms. Alternatively, the **EC$_{sos}$** , as an individual model, allows to be reduced from a superrecursive to a recursive algorithm. As it is clearly showed for ITM in [2], when the control device **A** comes in a final state from **F,** the inductive Turing machine **M** also stops functioning.

Once the pool of Lex A decreases, repression of the SOS genes goes down according to the LexA affinity to the SOS boxex. Operators that bind LexA weakly are the first to be fully expressed. In this way LexA can sequentially activate different mechanisms of repair. Genes having a weak LexA box (lexA, recA, uvrA,B,D) are fully induced in response to even weak SOS inducing treatments. Thus the first SOS repair mechanism to be induced is nucleotide excision repair (NER), whose aim is to fix DNA damage without commitment to a full-fledged SOS response.

Obviously, SOS repair covers wide range of self-maintenance capable for evolution hierarchy. The **EC$_{sos}$** model adequately introduces reparation of the first order toward inductive hierarchy that includes higher-orders of inductive computation of adaptation and evolvability with inductively defined memory. According to definition, the evolutionary memory **EM** is called inductive if the relation that provides connections between cells and all mappings of the structured memory are defined by some inductive Turing machine.[2] We could find the SOS repair and equivalenty the **EC$_{sos}$** model as basic foundation for developing innovative adaptation system within universal evolutionary computer represented in the following section.

We have seen that results appeared either with or without halting. That is in the case when recursive algorithms are sufficent and we could reduce superrecursive algorithms to recursive ones. In all other cases ITM continues with operation without stopping. It is possible construct the inductive Turing machine **G** such that **G** never stops and computes the same function as inductive Turing machine **M**, that is , **M** and **G** are functionally equivalent.[2] The **EC$_{hm}$** and **EC$_m$** model have been independently developed where each of them presents slightly different way of action (generally, targeted and untargeted mutations as well as hyper mut/rec functioning). However, they are functionally equivalent to **G**. The **EC$_{hm}$** allows us to simulate targeted mutations via equivalent functioning of hipermutable sequences. For example, the antibody producing lymphocytes of an immune system that mutate their hypervariable gene regions. The immune system processes without stopping, constantly trying to produce better solutions. Similarly, bacteria use much simpler tricks to target mutations to some specific genes, so-called contigency genes that are under strong selective pressure.[6] The **EC$_{hm}$** model could simulate this functioning in the direction of avoiding solutions equivalent to deleterious mutations and favouring only adaptive solutions. That relies on the fact that inductive computation of **EC$_{hm}$** , while working without halting, can ocassionally changes its output. The result could be good enough even if another (possibly better) result may arrive in the future. The structured memory **EM** allows the storing of adaptive solutions in the way as the immune system "memorizes" solutions through the mechanism of dividing and proliferation of the cells. So, the response of the immune system to the appearance of the same antigen is very quick.

There are situations when the result is already obtained but the inductive Turing machine cannot stop functioning. As a consequence the process of emergence becomes in some sense infinite although the result is always computed in a finite time. This feature of inductive computation arises within **EC$_m$** model. As a parallel example from the world of bacteria we could find bacterial ribosomal mutants resistent to high concentrations of streptomycin that have acquired very high ribosomal fidelity in the process of protein synthesis. Such mutants cannot grow any more without streptomycin addiction – the drug that increases the ribosomal error rate.[7]

Strong genetic mutators are particulary favored when adaptation requires several genomic mutations which may be the case of most adaptations to complex enviromental changes. Simulating this functioning via **EC$_m$** , the solutions of adaptive mutations could appear as an array or sequence of previously achieved results.[Table 1:6] Selecting for the mutator (which has been generated by inactivation of antimutator function) can allow fast exploration of the fitness landscape. Mechanistically, the mutator increases in its frequency because of its genomic association (hitch-hiking) with favorable mutations generated by mutators activity. This is supported in the models of inductive computations by structured memory. Time complexity reflects the speed of computations. It is possible to find that **EC$_m$** computes the same function as any recursive algorithm but with much greater speed and only if the function is sufficently complex. That is equivalent to harsh selective request of the complex changing and stresfull enviroment of bacterial mutants. In this case mutators have advantage over the inducible mutators because these ones may not have had time to produce adaptive solution.

The superrecursive algorithms of **EC$_{sos-mrs}$** model accelerate and increase the power and efficiency of evolutionary computation. The specific postreplicative mismatch repair mechanism (LPMR) prevents recombination (it edits anti-recombination "down" effect) between partially homologous sequences acting as effective genetic interspecies barrier. But when it has been coselected by mut/rec activity it becomes gradually deficient. With other words, when mismatch repair (MRS) is inactive (negative control off) and SOS system is turned on (positive control on) the genetic barrier is eliminated (in the experiment with Esherichia and Salmonella).[8] Hence, two related but different genomes could exchange and mosaicaly rearrange their genetic informations. The deficiency in the mismatch repair of two different bacterial species allows the exchange of genes that do not exist within individual species. The frequency of mutation and interspecies recombination rapidly increases as well as genetic diversity and consequently, probability of succesful adaptation. This biological genetic information processing exibits the power of adaptive computation of the higher order. The MRS ans SOS system appeared to be a homeostatic couple for the fine tuning of genetic variability: the former conserves genetic informations constituently, the latter increases its variability inductively. Such bacterial behaviour could be simulated by **EC$_{sos-mrs}$** model that is represented here as two different inductive Turing machines acting. The application of an inductive Turing machine to organization of the (genetic like) memory of another inductive Turing machine causes increase in computing power of machines. In some sense, the first one performs preprocessing of information for the second one. This is a nonlinear composition of inductive Turing machines which extends the computability space from one order to another many times.[2]

**Table 1.:Summary of resulting features of input, output and processing levels of independent models**

| Inductive Turing machine (smITM) | Evolutionary computer (ECsos; ECsos-mrs; EChm; ECm) |
|---|---|
| 1.Both model work with structured and finite objects | |
| 2.Processes are inductiv i.e.emergent | |
| 3.Results are obtained without stopping | |
| 4.Result emerges only in the corresponding computational process | |
| 5.Result appears as the last reached word that is not changing | |
| 6.Result appeared through the sequence of intermediate results | |
| 7.SR-algorithms accelerate and increase the power and efficiency | |

It turned out that two descriptions of algorithms generate the same sets of computational processes and they are pointed on the same type of action According to definition, if they compute the same class of functions or relations for nondeterministic algorithms we may conclude that they are equivalent.

Each of these presented systems, individually as well as mutually, are passing through three phases of evolutionary information processing: inducibility, proavtivity and specialization. Finally, it is possible to suggest the universal evolutionary computer that is able to efficiently simulate any evolutionary mechanism.

## 3. THE UNIVERSAL EVOLUTIONARY COMPUTER

The evolution is a huge theory and an immensely powerful creative process. But how huge and how powerful? The Old

Indian proverb says:"The world that could stay on the palm is just one of the many." If we consider the living Earth from its beging we find out that Life never stops. Every time when the evolutionary solvable problem related with survival and growth appeared, it has been solved by some evolutionary genom based computer. Even after the two most powerful extinction catastrophies, 250 and 65 million years ago [9], the Life has been continued and the species have been changed. One could doubt that the Life metagenom presented as universal genom computer, in all discrete moments from the very begging, with every possible change on its structure could possibly stay on the palm of the single being created by the genom computer itself. Actually, this whole figure is related only to the biological evolution. In addition, it is possible to abandoned the idea of the gen as the only base of our thoughts of evolution. Consequently, it is not necessary that storing and replicating structures have to be made of nucleic acids. The world of biological evolution based on genes could be only one among many of them. However, they have to be the self-maintaining information system capable for evolution. Adaptation to unpredictable changes requires flexibility and variability, whereas the maintenance of already evolved systems requires stability. That is possible to simulate by the universal evolutionary computer.

It is possible to invent (i.e. induce) a evolutionary computer which can be used to compute any computable sequence.
In every adaptive (or learning) system, some structures are undergoing adaptation. For nongenetic adaptive algorithms it is typically a single point in the search space of the problem. For conventional genetic algorithms or GP paradigm, such structures, are the individual points in a population of points from the search space of the problem. Generally, such or similar structures usually allow the selection only on the phenotype and the program organizations of the genome remain essentially unchanged throughout the evolution.
In the course of rigorous universal way of evolution this model introduces the basic processable unit of evolution-*epu*; (according to IPA phonetics: ipju) that is a neccessary eqvivalent for biological or any other replicators (e.g. genes, prions or hypothetical memes).
Evolutionary processable unit i.e. *epu* consists of two integrative major stuctures. The first one is programming or coding part. It includes coding sequence with its activator (or regulator) function and function of replication. The replication may be integrated in the coding sequence (by analogy, as replicase that catalyses its own replication without the help from a protein) or separately added (by analogy, coding for the product with its enzymatic actions).
The second part is not rigorously preprogrammed. It allows self-selective interactions and connections among the units of *epu* enviroment. Finally, this part also includes tendency towards *epu* allelic form of variations and functions of mutation and recombination. There is a wide range of *epu's* constructions but essentialy, any *epu*-construction inevitably remains within the definition of single $epu = \{f(s), f(r), f(i), f(v)\}$. The *epu*-constructable memory is *epuon*. It is equivalent to structured memory of inductive Turing machine.

Evolutionary *epu*-coding and replication via **S** computer, (where **S** is equivalent to a Turing machine **T**) allows development and connection with two integrative computing subsystems of evolutionary computer **EC**.[Fig.1.] Two sets of algorithms for evolutionary computation are $C_{ima}$ and $C_{ina}$ of imitative and innovative adaptive computation, respectively.
The first subsystem, $C_{ima}$, relies on replicative dinamycs. Solutions exist within *epu* homologous sequences that support processes of imitative adaptation. Under the repetitive selective pressure of the same or similar conditions on the input we could find that $C_{ima}$ computes the same function as **S**, that is, $C_{ima}$ and **S** are functionally equivalent. Additionally, $C_{ima}$ is functionally equivalent to **ITM** of the first order. Finally **EC** based on the $C_{ima}$ computation acts autonomously in the first degree. All standard functions from **S** have been codified through the codification process called *epuization,* so, there is no need for **S** functioning. The $C_{ima}$ may acts independently of S.
The second subsystem is inductive computation of inovative adaptation, $C_{ina}$, and its structure allows associative dinamycs. (That is, it accepts and performs instructions of the form «go to the epu i», or by analogy, it performs the «hitchhiking» form). The $C_{ina}$ includes wide range of repair systems of the first order (as nucleotide excision repair, NER, whose aim is to fix DNA damage without commitment to a full-fledged SOS response) through the second order repairs (actually adaptation), to the n-th order of self-accelerating evolvability (evolutionary adaptability) computation of indirect selection (systems that have «learnt to learn»), and finally toward n+1 order of entirely self-regenerating systems (as recently discovered mechanism of *D.Radiodurans* called extended synthesis-dependent strand annealing, ES-DSA, [10]).

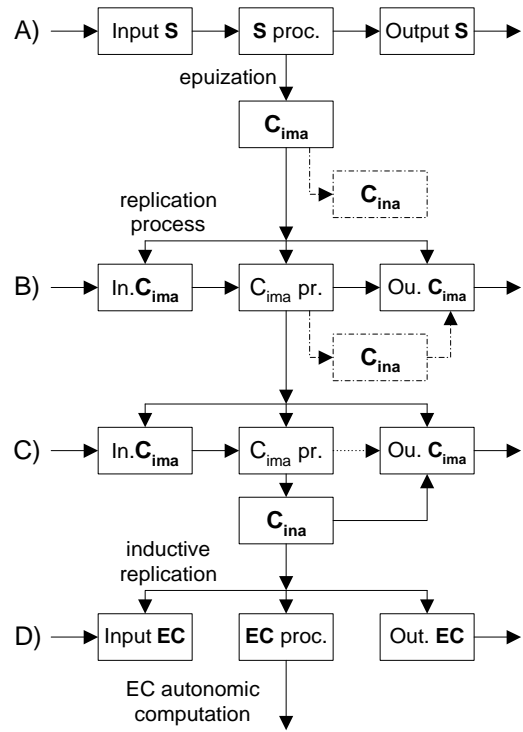

**Figure 1** : Simplified exposition of EC induction via triadic structure of information processing (see text)

We could recognize three phases of information processing of evolvability systems in general, and in the computation of $C_{ina}$ in particular: inducibility, proactivity and specialization. When the serious class of new problems (as a different sort of hard selective pressure) appears within partially homologous (i.e.homeologous) *epu* information sequences neither of the lower repair system can act efficiently based on insufficient *epu* knowledge base. So, the higher order repair systems will be induced gradually based on inductive proactivity mode (as it has been explained in the previous section). Specific resulting solutions are useful as a response to a single and reccurent selective input. In the complex enviromental changes emergent computational processes of adaptation require several specific mutations. Specialization, as the final phase of inductive computation, may include several specific useful *epu*-novelties. Actually, it evolves by changing and improving the procedures that are generating *epu* novelties themselves. Emerging processes of specific agents (specific enzymes by analogy) lead to instant evolution and completely new *epu*-sequences that have not been in preexisting library. The $C_{ina}$ computation (as ECsos, ECmrs, EChm) produces the definite result after the finite number of computing operations without stopping, that is, the final result emerges through the sequence of intermediate results. This computation effectively changes the memory structure of the system. While the epu's are not symbolic but realistic program entites the $C_{ina}$ computation acts autonomously in the second degree. That could possibly lead toward the systems of entirely autonomic computation and it can emerge almost exclusively on the non-deterministic basis through the corresponding computational processes.

## 4. IMPLICATIONS AND APPLICATIONS

The usual notion of solving complex problem is in the course of higher processing power and capacity. Yet, instead of "brute force" and exhaustive search through the space of all the alternatives, evolvability computation of superrecursive algorithms focus on improving program procedure that increases productivity and efficiency. Furthermore, there is no need for advanced forms of technologies for superrecursive evolvability computation. It can be induced by conventional Turing machines (by modification of recursive algorithms) and it can be commercially available before exotic technologies such as DNA or quantum computation. [2]

Theory of superrecursive algorithms enable us to see far beyond the grasp of practice. The evolvability computation based on the resourcefull evolution with number of computable models and strategies can help us to bridge the existing gap between the theory and practice. Several evolvable information sytems are already in progress.

The complex data-driven decision support systems are usually based on the datawarehouse and on-line analitical proccesing tools. Evolvability computation allows development directly from the level of transaction information systems and databases. The development emerges as the custom suitable applications from interactions between human users and evolvability computer.The simulation shows that single primordial model could evolve into the number of small offspring apps. Each of them could be highly specialized for specific user tasks (marketing, finance etc.) but they are still working interdependently as a whole. This opens new doors for applications in the organizational intelligence and specifically for ubiquitous and pervasive intelligence management and CRM. Traditional reporting systems become obsolete.

After we have taken knowledge from several remarkable lessons of evolution, now we can go back to biology, biotechnology and experimental evolution with a new tool. Potential applications appear almost limitless.

UEC conjugates computing and biology simultaneously for the benefit to both: evolution emphasises realization of superrecursive computation paradigm as a new way of computing, opens possibilities for a human-computer cooperation to an unimaginable degree and hence, catalyzes through this channel a new way of understanding the living world including the details of how evolution works. As Langton's intriguing hint says:"Life isn't just *like* a computation, in the sense of being a property of the organization rather than the molecules. Life literally is a computation"